\documentclass{article} %
\usepackage{iclr2027_conference,times}

\usepackage{amsmath,amsfonts,bm}

\def\eqref#1{equation~\ref{#1}}
\def\1{\bm{1}}

\DeclareMathAlphabet{\mathsfit}{\encodingdefault}{\sfdefault}{m}{sl}
\SetMathAlphabet{\mathsfit}{bold}{\encodingdefault}{\sfdefault}{bx}{n}

\usepackage{hyperref}
\usepackage{url}
\usepackage{booktabs, tabularx, amsmath}
\usepackage{graphicx}
\usepackage{subcaption}
\usepackage{multirow}
\usepackage{makecell}
\usepackage{xcolor}
\usepackage{wrapfig}
\usepackage{float}

\title{Don't Skip the Baseline:\\A Controlled Comparison of Dynamic\\Expert Skipping in MoEs}

\author{%
  Jakob Steimle$^{1,2}$\quad
  Thomas Elsken$^{2}$\quad
  Daniel Ehrhardt$^{2}$ \\
  \textbf{Lukas Rinder$^{2}$\quad
  Daniel Cremers$^{1}$\quad
  Xi Wang$^{1,3,4}$} \\[0.5ex]
  $^{1}$Technical University of Munich\quad
  $^{2}$Tensordyne\quad
  $^{3}$ETH Zurich\quad
  $^{4}$Ulm University \\
  Correspondence: \texttt{jakob.steimle@tensordyne.ai}
}

\usepackage{xspace}
\usepackage{placeins}

\newcommand{\EK}{$K_\text{eff}$\xspace}
\newcommand{\EKn}{$K_\text{eff}^\text{calibrated}$\xspace}
\newcommand{\topk}{top-$k$\xspace}

\iclrfinalcopy

\hypersetup{
  pdftitle={Don't Skip the Baseline: A Controlled Comparison of Dynamic Expert Skipping in MoEs},
  pdfauthor={Jakob Steimle, Thomas Elsken, Daniel Ehrhardt, Lukas Rinder, Daniel Cremers, Xi Wang}
}

\begin{document}

\maketitle
\fancyhead{}
\lhead{Preprint}   %
\renewcommand{\headrulewidth}{1pt}

\begin{abstract}
A growing body of work proposes \emph{dynamic expert-skipping} methods for Mixture-of-Experts (MoE) models that adjust the number of experts activated per token during inference, aiming to preserve model quality while further reducing the number of active parameters. These methods have added calibrated importance scores, inter-expert similarity metrics, or learned per-token allocators in order to identify experts to skip. However, different calibration regimes, different reductions in the effective number of experts, and evaluation suites make comparisons difficult. We classify four representative prior methods into a unified taxonomy and benchmark them against a natural, simple baseline:  statically reducing the number of experts $k$ (i.e., same reduced $k$ across all layers and tokens).  We provide a fair comparison of these methods at matched expert-compute budget, on two modern MoEs (Gemma4-26B-A4B and Qwen3-30B-A3B) in a text-only setting and a contemporary vision-language model (Qwen3-VL) in a multimodal setting. Surprisingly, we find that the simple baseline is hard to outperform consistently across models, modalities, benchmarks, and calibration settings. The methods we benchmark often underperform the baseline and achieved gains, when present, are minor. We further find that the effective $k$ achieved (or optimized for) in calibration can diverge from the effective $k$ in testing, resulting in over- or under-estimating the compute at inference time. Based on these findings, we recommend a calibration-free per-modality static $k$ reduction and validate this on a held-out model and modality. We further argue that a static control at matched \textit{measured} compute is the baseline any dynamic skipping method must beat.

\end{abstract}

\section{Introduction}
Mixture-of-Experts (MoE) models have become one of the leading paradigms for model architectures, activating only a small subset of all parameters (called ``experts'') during each forward pass. %
Typical models~\citep{yang_qwen3_2025,deepseek-ai_deepseek-v3_2025, team_gemma_2026} activate $k=8$ out of 128 to 512 available experts---fixed across all layers and tokens. Some methods for selecting a reduced number of experts already during (pre)training have been proposed \citep{raposo_mixture--depths_2024, zhou_mixture--experts_2022, gao_anyexperts_2025, jin_sparsity-controllable_2025, guo_dynamic_2025}, however these approaches are not yet adopted in major open-weight model families.
We therefore focus on the question: given a trained model, can we reduce the active number of experts while preserving quality by reducing this $k$ during inference, potentially dynamically per token and/or per layer?

Static, fixed-$k$ MoE inference has clear practical advantages: it is simple to implement, requires no additional calibration (i.e., selecting $k$ given an input token and layer), and, most importantly, matches the routing behavior learned during training. It further produces predictable, uniform compute and memory-access patterns across tokens. However, research has shown that not every token requires the same number of experts~\citep{lu_not_2024}. As a result, different approaches have been proposed to optimize inference costs by varying $k$. These approaches include static pruning, which permanently reduces the set of available experts to conserve memory, and (dynamic) skipping, which instead varies the number of active experts per token at inference time. \citet{yang_faster_2025} describe three benefits dynamic skipping can offer over fixed-$k$ inference: first, it reduces the total number of computations performed during each forward pass. Second, in small-batch scenarios, it reduces the number of unique experts active, lowering memory bandwidth requirements. Third, in a distributed deployment scenario, it can reduce communication overhead and improve load balancing.

Some recent methods have considered dynamic expert skipping as an add-on to static model pruning, rather than evaluating its benefits on its own, making it difficult to disentangle how much of the reported benefit comes from skipping itself versus the accompanying pruning. Beyond that, the proposed methods have become increasingly complex and costly. These range from calibrated scores~\citep{lu_not_2024} to similarity approaches~\citep{bai_diep_2025}, and learned allocators~\citep{yue_ada-k_2025}. One of the most recent approaches by~\citet{huang_modes_2026} is MoDES, which uses calibrated modality-aware thresholds and reports near original performance while reducing the number of active experts by over $80\%$. The combination of different calibration regimes, operating points (method configurations), and evaluations makes comparison difficult.
To close this gap, this study aims to provide a fair comparison across methods, benchmarks and inference-compute budget.

Our contributions:
\begin{enumerate}
   \item A unified taxonomy of dynamic expert-skipping strategies, organized by the signal
   they add over the standard MoE router (Table~\ref{tab:taxonomy_minimal}).
   \item A controlled, matched expert compute comparison of four representative methods against a
   static \topk baseline across different modalities and models, made fair by measuring compute at test time and comparing the full quality-compute frontier. 
   \item A zero-calibration, per-modality static \topk{}
configuration ($k_\text{vision}=1$--$2$, $k_\text{text}=6$) that recovers
$\approx95\%$ of original quality at $\approx 85\%$ fewer active experts in vision-heavy tasks.
   \item The insight that, despite their added complexity and cost, contemporary dynamic skipping methods do not systematically outperform a simple static baseline at matched compute.
\end{enumerate}

We find that, across models, modalities, benchmarks, and calibration settings, the simple baseline is hard to outperform consistently and by large margins. To allow reproducibility we will release our code, including \mbox{(re-)implementations} of all evaluated skipping methods. This further allows the original authors to verify our implementations.

\section{Background}
A number of methods have been proposed to leverage the different expert needs of tokens. We distinguish between \emph{expert skipping}, which reduces the set of experts activated per token, and \emph{expert rerouting}, which selects different experts than those chosen by the router.

\paragraph{Expert Skipping}
methods target existing static top-$k$ models. An early such strategy, ``Not all Experts are Equal'' (NAEE) \citep{lu_not_2024},
skips a tail expert whenever its routing score falls short of the top expert's
by a calibrated margin; \citet{bai_diep_2025} extend this criterion with an
inter-expert redundancy term. Further criteria include per-token sensitivity
\citep{zhong_adapmoe_2024}, confidence-triggered re-evaluation with the full
expert set \citep{zhao_mobile_2025}, and data-free per-layer expert budgets
\citep{chitty-venkata_lexi_2025}. \citet{huang_modes_2026} extend
threshold-based skipping to the multimodal domain with per-layer importance
scores and per-modality thresholds, while \citet{yue_ada-k_2025} train a
per-token classifier via reinforcement learning to predict the required expert
count.

\paragraph{Expert Rerouting}
methods alter \emph{which} experts are selected rather than how many,
steering tokens toward cached or already-active experts to reduce
weight-loading overhead \citep{skliar_mixture_2024, wang_buddymoe_2025,
wu_sere_2026, oncescu_opportunistic_2025}. Their benefit hinges on shrinking
the unique set of active experts per forward pass. This reduces overhead due to memory bandwidth, but only works in small batch sizes. At sufficiently large batch sizes nearly all experts are required.

For this paper we focus on dynamic skipping, as rerouting offers limited benefit in common (large batch) inference scenarios. We explicitly study skipping in isolation from the pruning stages that some of these methods originally include. As skipping is orthogonal and composable with any pruning method, this lets us attribute quality differences to the skipping decision itself. Closest to our baseline (next section), \citet{xia_fastmmoe_2025} reduce the expert
count for vision tokens to a fixed $k_\text{vision} < k$ from a chosen
onset layer onward, while text tokens always retain the full $k$; they
ablate ($k_\text{vision}$, onset layer) at moderate reductions
($k_\text{vision} = 4$--$5$ of $8$) and deploy the mechanism jointly
with visual token pruning.

\section{Methodology}
\subsection{Baselines}
\label{sec:baselines}
Our natural baseline for expert skipping methods builds on the model's native routing: statically reducing the number of experts $k$. We refer to this baseline as softmax \topk, based on the selection criteria. As a reminder, for popular mixture of expert models like Qwen3, the router generates logits, which are normalized with softmax before the \topk experts are selected. We expand on this approach for multimodal models by defining separate $k$ values for each modality: $k_\text{text}$ and $k_\text{vision}$. This separate treatment of modalities is inspired by prior work \citep{huang_modes_2026,xia_fastmmoe_2025}. For models that support image and video tokens separately, we use $k_\text{vision}$ for both token types. Both baseline variants share the same benefit: they are calibration-free and static.

\subsection{Methods for comparison}
\label{sec:methods}
\begin{table}[t]
\centering
\caption{Taxonomy of expert-skipping strategies, ordered by increasing signal complexity over the standard router. Calibration time is per operating point, measured on a single NVIDIA A100 80GB on Qwen3/Qwen3-VL and ranges from 30 minutes to 20 hours depending on the method. Full taxonomy in Appendix~\ref{app:taxonomy}.}
\footnotesize
\setlength{\tabcolsep}{4pt}
\begin{tabular}{@{}llllc@{}}
\toprule
\textbf{Method} & \textbf{Criteria} & \textbf{Decision} & \textbf{Tunable} & \textbf{Calib. (h)} \\
\midrule
Softmax top-$k$ (baseline) & Gate weights & Keep top-$k$ & $k$ & --- \\
Per-modality (ours) & Gate weights & Top-$k$ per modality & $k_{\text{text}}$, $k_{\text{vision}}$ & --- \\
\midrule
NAEE & Gate weights (cum.) & Cumulative-mass threshold & $\beta$ quantile & 0.5 \\
DiEP & Gate weights + CKA & Redundancy-scaled ratio & $\gamma_1$ quantile & 0.5 \\
MoDES & Gate prob.\ $\times$ layer imp. & Per-modality threshold & target skip ratio & 20 \\
Ada-K & Hidden state (learned) & Predicts expert count & $\lambda$ & 2.5 \\
\bottomrule
\end{tabular}

\label{tab:taxonomy_minimal}
\end{table}

We evaluate four representative dynamic skipping methods:
NAEE~\citep{lu_not_2024}, DiEP~\citep{bai_diep_2025}, MoDES~\citep{huang_modes_2026}, and Ada-K~\citep{yue_ada-k_2025}, summarized in Table~\ref{tab:taxonomy_minimal}. We describe them based on a number of concrete properties of the methods: their criteria, decision rule, calibration phase and tunable parameters. We further discuss the calibration time necessary for a single operating point.
Both NAEE and DiEP also provide methods for expert pruning that they combine with the skipping approach. We only consider the dynamic skipping approach here.

\paragraph{Skipping Criteria} The first major differentiating factor between the methods is the skipping criteria. While our baselines and NAEE skip experts based entirely on the routing scores (also called gate weights), the other methods expand on this: DiEP uses an inter-expert Centered Kernel Alignment (CKA)~\citep{kornblith_similarity_2019} statistic that measures redundancy in individual experts, while MoDES adds a global layer importance factor. Ada-K is different here, as it uses the hidden state of the token rather than the routing scores as a criteria.

\paragraph{Decision Rule} Our baselines have the simplest rules, as they retain the \topk experts and vary only the $k$. NAEE on the other hand uses the cumulative routing scores and compares their ratio against a calibrated threshold. DiEP and MoDES adapt the concept from NAEE with a more complex threshold and score. All methods except for DiEP always keep the expert order intact. Further, all dynamic methods except for MoDES will always retain at least the first selected expert. Ada-K differs from the other methods in its decision rule: rather than deciding whether a given expert should be skipped, it instead predicts the number of experts that should be retained using what they call an \textit{Allocator}.

\paragraph{Calibration} While our baselines are calibration free, NAEE, DiEP and MoDES all require calibration. NAEE calibrates the threshold value used for skipping, while DiEP calibrates this ($\gamma_1$) as well as the CKA similarity metric between experts ($\gamma_2(e_0,e_i)$). MoDES first calibrates the importance $\alpha^{(l)}$ of each layer and then identifies the ideal thresholds ($\tau_{v,t}$) needed to achieve a target skipping ratio. Ada-K uses a unique approach, as it optimizes the linear layer of the Allocator using Proximal Policy Optimization (PPO)~\citep{schulman_proximal_2017}, a type of reinforcement learning.

\paragraph{Tunable Parameters} Tunable parameters can be adjusted to vary the targeted operating point. For NAEE this comes in the form of the $\beta$ quantile. When calibrating, the cumulative ratio between the routing scores is collected for each layer. By selecting an appropriate quantile we can sweep across operating points, skipping more or less experts. NAEE natively uses only the median. For DiEP we again sweep the quantile for the skipping parameter $\gamma_1$, in their paper they primarily use the median as an operating point. For MoDES the tunable parameter is the target skip ratio, as part of their calibration procedure they find the ideal thresholds ($\tau_{v,t}$) that achieve the target skip ratio. Lastly Ada-K has one main parameter $\lambda$. This defines the cost of adding additional experts in the strategy. A higher cost $\lambda$ results in fewer experts being predicted to be needed.

\paragraph{Calibration Cost} We calculate the calibration cost using the original code release for MoDES and use our own implementation for the other methods on a single A100 80GB. While our static baselines are free, since they are calibration free, the advanced dynamic skipping methods have significantly different requirements. DiEP and NAEE are both relatively fast to compute, requiring approximately half an hour to calibrate for Qwen3-30B-A3B, depending on calibration data. Ada-K also varies significantly, but uses a much larger training dataset, resulting in around 2.5 hours of training time. MoDES requires the longest calibration at nearly 20 hours, as both the layer importance score measurement and the frontier search involve many forward passes through the full model.

We selected these methods as representative: they are current and use significantly different approaches to skipping. DiEP with its consideration of expert redundancy, Ada-K with its training approach, MoDES due to the multimodality and NAEE as the simplest calibration-based approach all reflect distinct criteria for selecting experts that can be skipped. 

\subsection{Quantifying Per-Token Expert Effort}
\label{sec:effective_k}
All considered skipping strategies activate fewer experts for some tokens. To compare these savings on a common, hardware-agnostic axis, we define effective K (\EK) as a proxy metric. \EK is the mean number of routed experts per token, averaged across all tokens and layers: $K_{\text{eff}} \;=\; \frac{1}{T\cdot L}\sum_{t=1}^{T}\sum_{l=1}^{L} k_{t,l}$,
where $k_{t,l}$ is the number of experts activated for token $t$ in layer $l$. For the static softmax \topk baseline, $K_\text{eff} = k$ by construction, whereas for the dynamic strategies, \EK is a measured average that depends on the inputs and the strategy's configuration. We distinguish between \textit{calibrated} \EKn and \textit{measured} \EK. \EKn is based on the number of active experts for the calibration set~\citep{huang_modes_2026}. Our analysis uses \textit{measured} \EK, which counts expert activations at test time during the quality evaluation. The two may diverge because any threshold only enforces the target value on the calibration set, but may transfer imperfectly across datasets and depends heavily on the modality split, as demonstrated in Section~\ref{sec:nominal-and-measured}.

\begin{figure}
    \centering
    \includegraphics[width=0.9\linewidth]{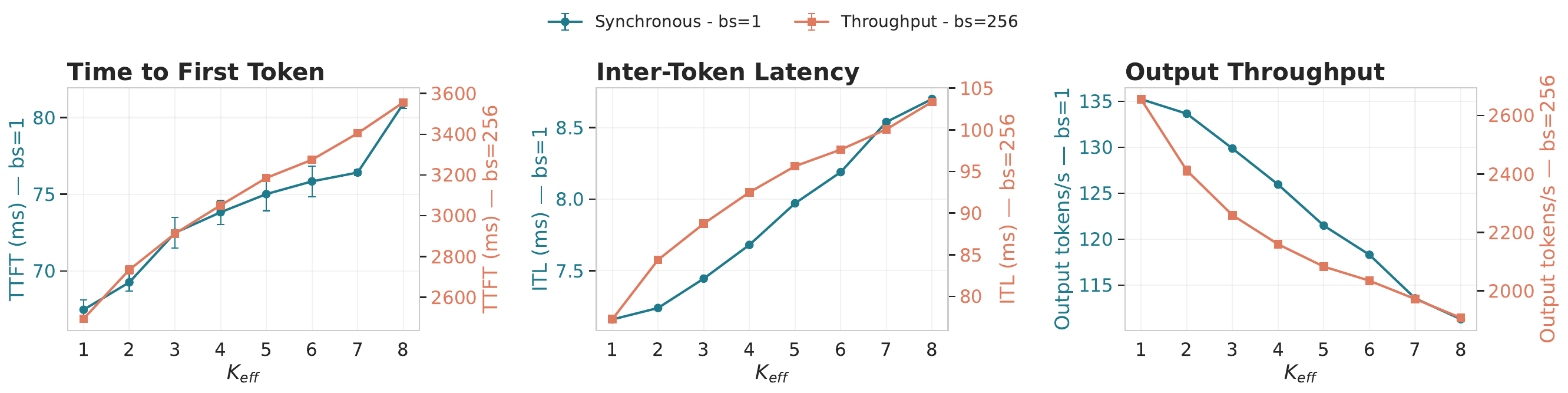}
    \caption{Static \topk override, $k\in\{1,\dots,8\}$, vLLM on a single A100 80GB, GuideLLM chat scenario, batch sizes 1 (bandwidth-bound) and 256 (compute-bound); average across a 120 second load. TTFT, ITL and output throughput scale monotonically with the number of experts.}
    \label{fig:eff_k_vs_perf}
\end{figure}

The bottleneck for MoE inference depends on the deployment regime. At sufficiently large batch sizes, the expert feed-forward evaluation becomes compute-bound. In this case reducing the number of FLOPs provides the largest inference speed-up. The FLOPs of the expert matmuls scale linearly with the number of active experts per token: \EK is thus directly proportional to expert compute. Reducing active experts also lowers secondary costs such as all-to-all communication as described by~\citet{yang_faster_2025}. \EK counts only expert activations, ignoring other factors like routing, load imbalance, and communication overhead. This makes it a hardware-agnostic proxy for runtime cost, with true cost larger than what \EK implies. Figure~\ref{fig:eff_k_vs_perf} shows the monotonic but sublinear relationship between \EK and core inference metrics; we discuss this relation in Appendix~\ref{append:runtime_performance}.

\section{Results}
\subsection{Implementation}
Several of these methods provide incomplete or no code releases, so we reimplement them on a best-effort basis. Ada-K has no release at all, and we reproduce its PPO-trained classifier ourselves (Appendix~\ref{append:Allocator_training}), referring to it as the Allocator throughout. NAEE and DiEP release code only for Mixtral, a coarse-grained model, so we extend both to the $k=8$ regime. Where a method natively exposes only a single operating point, we sweep its free parameter: the quantiles from $0.1$ to $0.9$ for NAEE and DiEP, thresholds for MoDES in order to trace a full frontier. We reproduce reported operating points from each original paper to confirm our implementations are faithful (Appendix~\ref{append:validating_implementations}).
\subsection{Text-only Evaluation}
We compare all strategies at matched \EK on Qwen3-30B-A3B
\citep{yang_qwen3_2025} and Gemma4-26B-A4B \citep{team_gemma_2026}, two popular
mid-sized MoE models with $k=8$ active experts. While most methods report only a single
operating point (i.e., a single \EK) at which they report performance, we sweep each method's
tunable parameters to trace a full quality--compute frontier, providing clearer evidence
of each method's benefit at any given compute level. We evaluate on the
zero-shot multiple-choice benchmarks HellaSwag \citep{zellers_hellaswag_2019}
and ARC-C \citep{clark_think_2018}, as well as 5-shot GSM8K \citep{cobbe_training_2021} and zero-shot HumanEval~\citep{chen_evaluating_2021}
as generative benchmarks. A full overview of the benchmarks and corresponding
calibration sets is given in Table~\ref{tab:benchmarks}. We use
lm-eval~\citep{eval-harness} for
consistency with related literature. \EK is measured at test time during each evaluation (Section~\ref{sec:effective_k}); for GSM8K and HumanEval this includes
the autoregressive decoding passes. Since most methods require calibration, we define a \emph{realistic} calibration dataset, consisting of a mixed corpus of English, Chinese, code,
and math built from \textit{C4\_en} and \textit{C4\_zh} \citep{raffel_exploring_2023},
\textit{open-web-math} \citep{paster_openwebmath_2023}, and
\textit{python-codes-25k} \citep{flytech_python_codes_25k}. This realistic regime
simulates the calibration--evaluation mismatch typical of actual deployments. We also evaluate the impact of calibration by evaluating a ``best-case'' scenario on the benchmarks' own train sets.

\subsubsection{Strategy comparison}
\label{subsec:realistic-case}
\begin{figure}[t]
    \centering
    \begin{subfigure}[b]{0.48\linewidth}
        \centering
        \includegraphics[width=\linewidth]{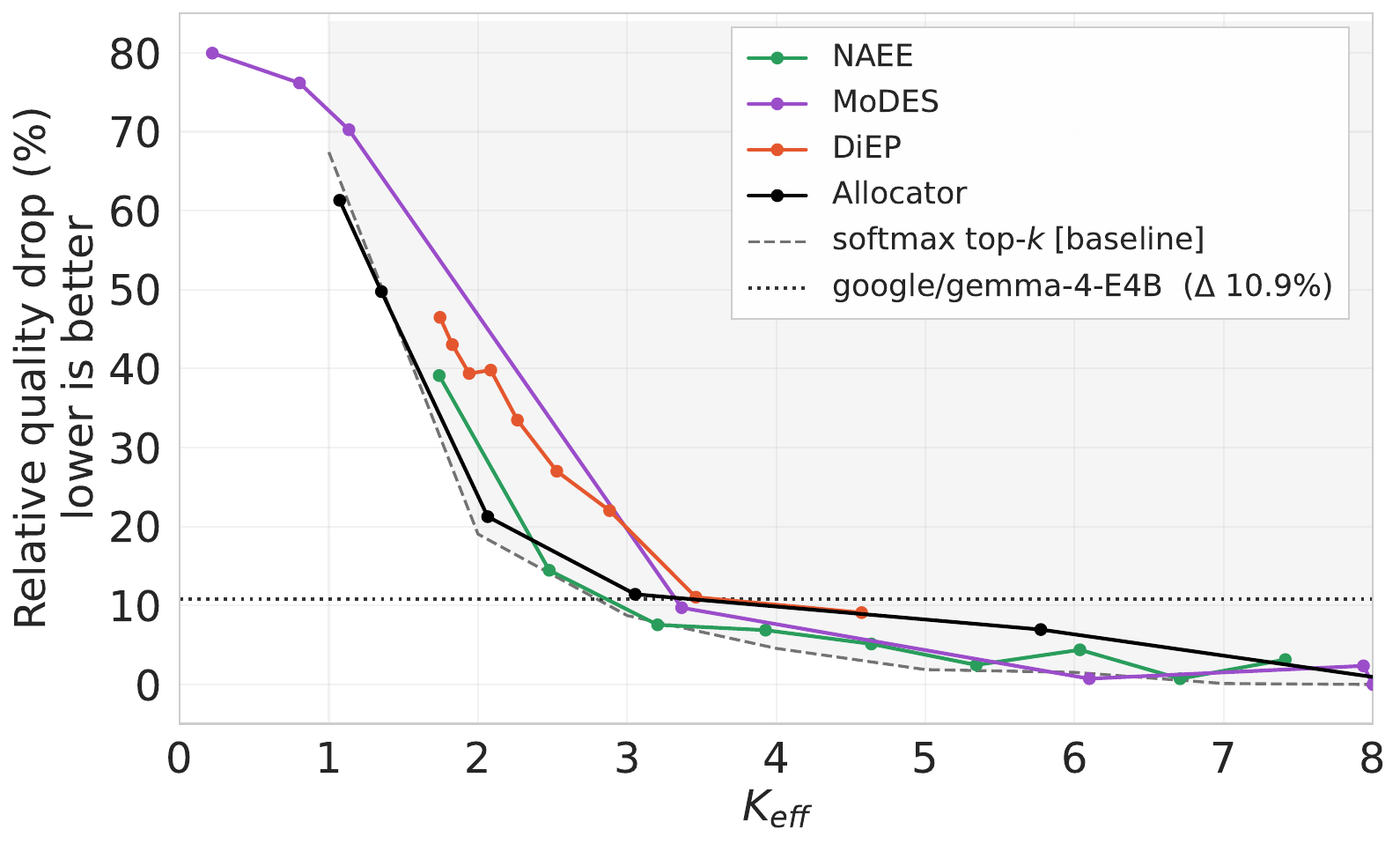}
        \caption{Qwen3-30B-A3B}
        \label{fig:relative_performance_drop_realistic_case_qwen}
    \end{subfigure}\hfill
    \begin{subfigure}[b]{0.48\linewidth}
        \centering
        \includegraphics[width=\linewidth]{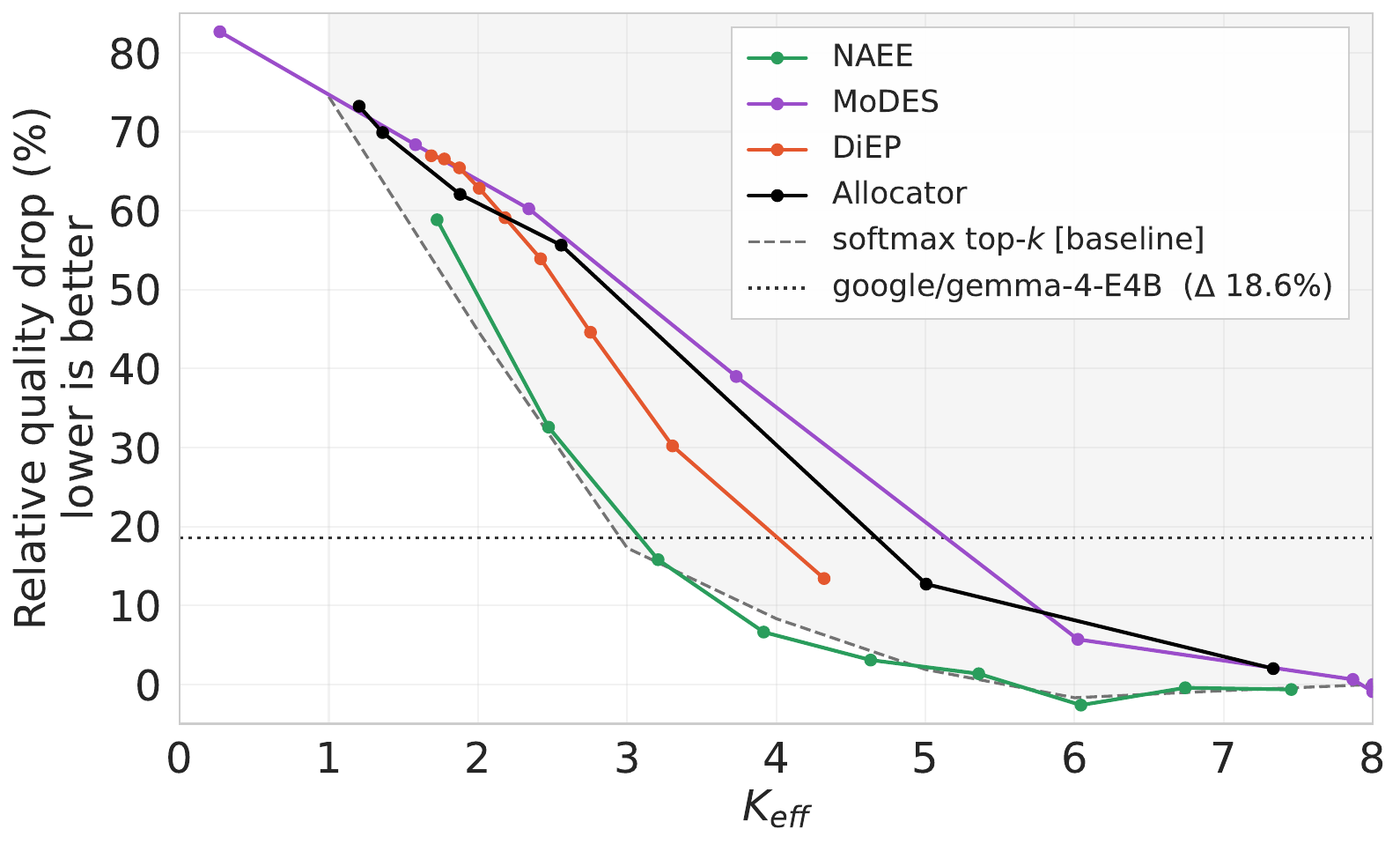}
        \caption{Gemma4-26B-A4B}
        \label{fig:relative_performance_drop_realistic_case_gemma}
    \end{subfigure}
    \caption{Relative performance drop (average across data sets) for different \EK values with strategies calibrated on a mixed dataset. We include the smaller Gemma4-E4B as a reference. The shaded area highlights performance below the baseline. No advanced skipping method systematically outperforms the baseline. Below \EK$\approx3$ it is preferable to use a smaller model.}
    \label{fig:relative_performance_drop_realistic_case}
\end{figure}

Figure~\ref{fig:relative_performance_drop_realistic_case} shows the difference in performance between the strategies introduced in Table~\ref{tab:taxonomy_minimal}. The plot compares the mean relative performance drop across ARC-C, HellaSwag, HumanEval and GSM8K for different \EK. The Pareto frontier reveals a number of behaviors: first, NAEE most closely matches the fixed \topk{} baseline for most points along the frontier, with some minor improvements compared to the baseline. The Allocator also shows some improvements in the \EK$\approx1$ regime for Qwen3. However, overall benefit is very limited, at most $\sim2.6$\,pp compared to the interpolated baseline. This difference is especially small given the substantial quality loss already occurring in this \EK range: at this point the model has already lost 50\% to 65\% of its original capability. To provide context, at \EK$\leq2.5$ the best strategy on Gemma4-26B-A4B performs worse on average than Gemma4-E4B, a dense model with $85\%$ fewer effective parameters\footnote{Following the Gemma4 model card~\citep{team_gemma_2026}, the count excludes embeddings; the 26B designation of the MoE follows the same convention.}. This limits practical use of the observed gains. Both MoDES and DiEP mostly perform worse than the baseline, with MoDES only slightly ahead above \EK$=1$. MoDES does, however, achieve operating points below \EK$=1$, meaning that for some tokens the MoE block is completely skipped, a regime that the baseline cannot reach.

\begin{figure}
    \centering
    \includegraphics[width=\linewidth]{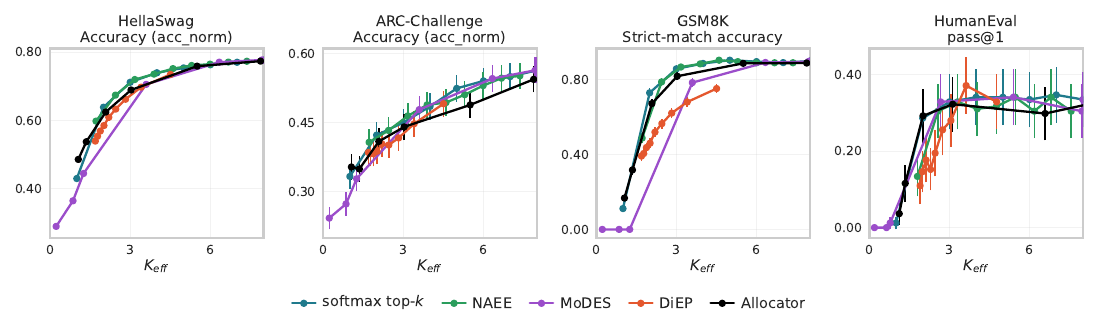}
    \caption{Absolute quality results of the strategies across four benchmarks on Qwen3. Multiple-choice benchmarks collapse smoothly towards random selection, generative tasks display clear collapse for small~$k$.}
    \label{fig:absolute-quality-text}
\end{figure}
\begin{figure}[!h]
    \centering
    \includegraphics[width=\linewidth]{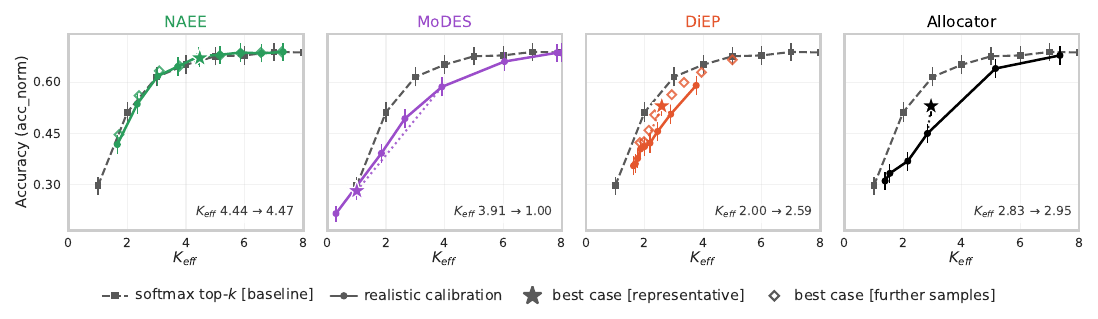}
    \caption{Shift in \EK and quality when using best-case calibration rather than realistic calibration on ARC-C on Gemma4. Best case calibration can help bridge the gap to the baseline but does not change the overall pattern.}
    \label{fig:text_calibration_minimal}
\end{figure}

The composition of the drops also matters: as Figure~\ref{fig:absolute-quality-text}
shows, the multiple-choice benchmarks degrade smoothly toward their
random-guessing floors, while GSM8K collapses sharply once \EK{} falls
below a model-dependent threshold (${\sim}2$ for Qwen3). HumanEval follows a similar pattern. Aggregate drops at low \EK{} are therefore
dominated by generative-task failure, and the safe operating point is
set by the generative benchmark for each model, not the average. Detailed per-benchmark results for Gemma4 are included in Appendix~\ref{append:benchmark_results}.

To validate that this is not an artifact of our realistic calibration, we also evaluate a best-case scenario using the benchmarks' own trainset. 
Figure~\ref{fig:text_calibration_minimal} highlights the shift of the middle operating point when using best-case calibration. For DiEP and MoDES a substantial shift in \EK can be observed and both DiEP and the Allocator gain substantially compared to the realistic calibration. However, this gain does not enable the methods to outperform the static baseline. We provide further details in Appendix~\ref{app:calibration_sensitivity}.

\FloatBarrier

\subsection{Multimodal Evaluation}
\subsubsection{Setup}

\begin{wrapfigure}{}{0.6\textwidth}
    \centering
    \includegraphics[width=\linewidth]{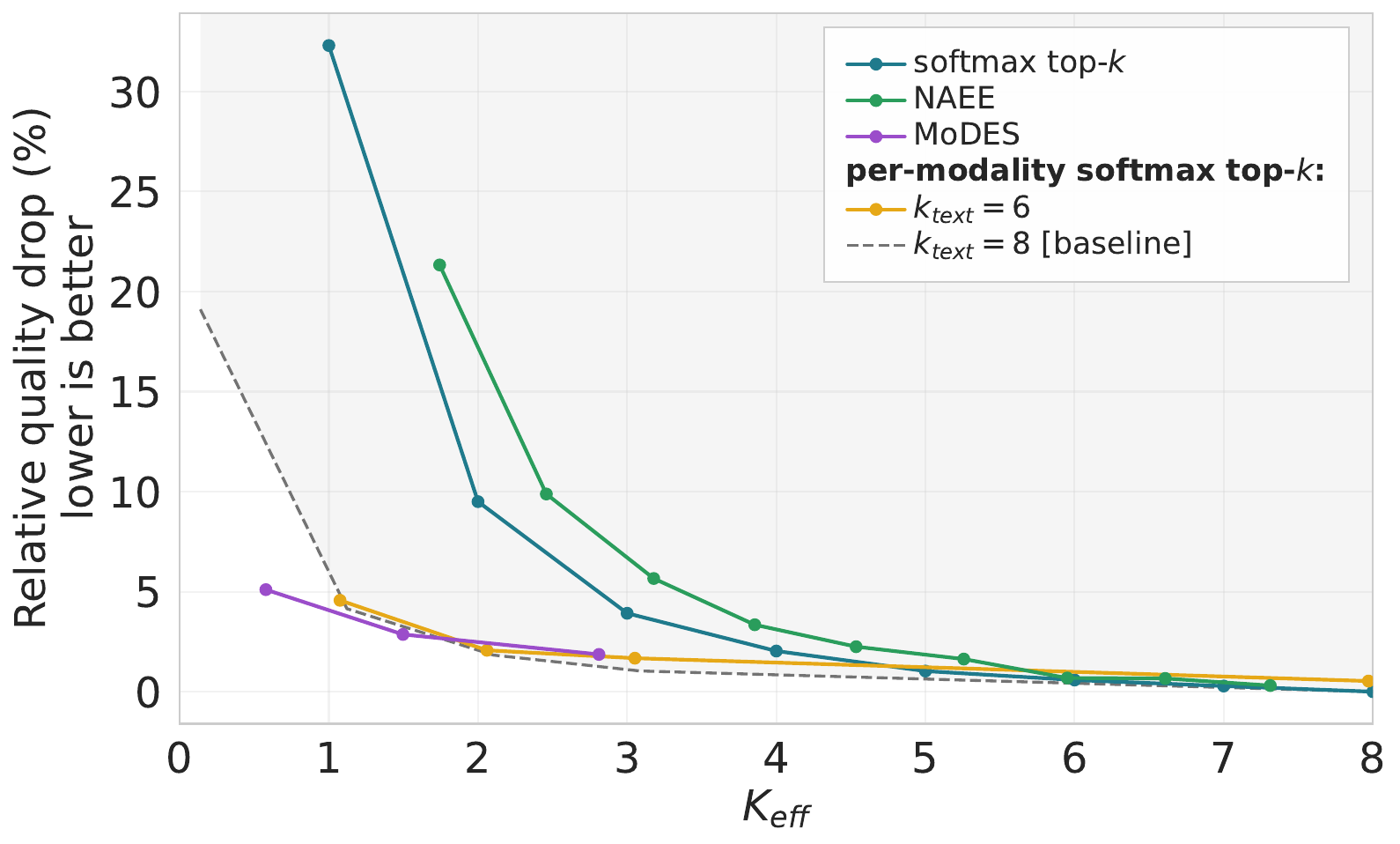}
    \caption{Aggregated relative quality drop for different \EK values with multimodal strategies. The horizontal dashed line refers to the performance of the original model. The dashed curve/shaded area highlights the baseline that only reduces $k_\text{vision}$. Modality awareness provides an immediate benefit to both our baseline and MoDES.}
    \label{fig:quality_drop_multimodal}
\end{wrapfigure}

In the multimodal setting, we compare NAEE and MoDES against the static softmax
\topk{} baseline and our per-modality variant. The per-modality variant is motivated by prior observations that vision tokens tolerate substantially fewer experts than text tokens \citep{xia_fastmmoe_2025, huang_modes_2026}. DiEP, which underperformed the
baseline in both text regimes, is excluded. The Ada-K Allocator is excluded as well. This is due to its loss formulation: the Allocator uses next-token prediction as its quality measure. Due to the training procedure of vision-language models, the per-token loss of vision tokens is never used during training, so they do not provide a quality signal that the Allocator could leverage. Applying the Allocator to multi-modal evaluations requires the implementation of a new loss formulation. As MoDES is natively designed for multimodal models, we expect
it to perform best in this scenario. We use Qwen3-VL-30B-A3B-Instruct
\citep{bai_qwen3-vl_2025}, closely related to the Qwen3 model used in our text-only
evaluation. To provide ideal conditions for MoDES, we calibrate it using its
published GQA \citep{hudson_gqa_2019} protocol; for consistency, we also calibrate NAEE
on GQA\@. We evaluate on ChartQA \citep{masry_chartqa_2022},
TextVQA \citep{singh_towards_2019}, and MME \citep{fu_mme_2025}, as well as DocVQA \citep{mathew_docvqa_2021}, InfoVQA \citep{mathew_infographicvqa_2021}, and OCRBench \citep{liu_ocrbench_2024} for text-heavy images. As in the
text-only setting, we measure expert activations during evaluation to obtain
\EK{}, computed over all routed tokens regardless of modality; we
report the per-benchmark modality mix in Appendix~\ref{append:benchmarks}.
Evaluation uses LMMS-Eval~\citep{bo_li_lmms-eval_2024,zhang_lmms-eval_2024}.

\subsubsection{Multimodal Results}

Figure~\ref{fig:quality_drop_multimodal} shows the relative drop compared to the
full model across all six
benchmarks, highlighting the performance differences across strategies.
MoDES, which is modality-aware, outperforms NAEE as well as softmax \topk{} by a large margin. However, it fails to outperform the per-modality softmax \topk{} baseline. Again, MoDES finds \EK$<1$, which is not reachable by the baseline except when skipping the experts for vision tokens completely.

The only outlier is MME-Cognition, a sub-task of MME,
(Figure~\ref{fig:absolute_metrics_multimodal_stderr}, Appendix), where MoDES
consistently leads the other approaches. The per-modality softmax strategy
provides two recommended trade-off points: for
$k_\text{text}=8$, reducing $k_\text{vision}$ to 1 incurs a drop
of approximately $4.1\%$, while reducing
$k_\text{vision}$ to 2 incurs an average drop of less than
$1.8\%$. Reducing $k_\text{vision}$ to 1 reduces the
number of active experts during the benchmark by
$\approx85\%$, to a
measured \EK{} of 1.08 on average. Reducing $k_\text{vision}$
to zero increases the drop to $19.1\%$. Surprisingly, this means
that more than $80\%$ of quality is retained even though the vision
tokens skip the experts entirely. Reducing $k_\text{text}$ to six allows further optimization of
the number of active experts, while incurring a small quality drop compared to
$k_\text{text}=8$. 
The benefit of modality-awareness is direct: reducing experts to one for
\emph{both} modalities (softmax \topk) costs more than $30\%$ of overall performance, while
even the more extreme $k_\text{vision}=0$ with text untouched (per-modality softmax \topk) costs only
$19.1\%$.

\begin{wrapfigure}{}{0.5\textwidth}
    \centering
    \includegraphics[width=\linewidth]{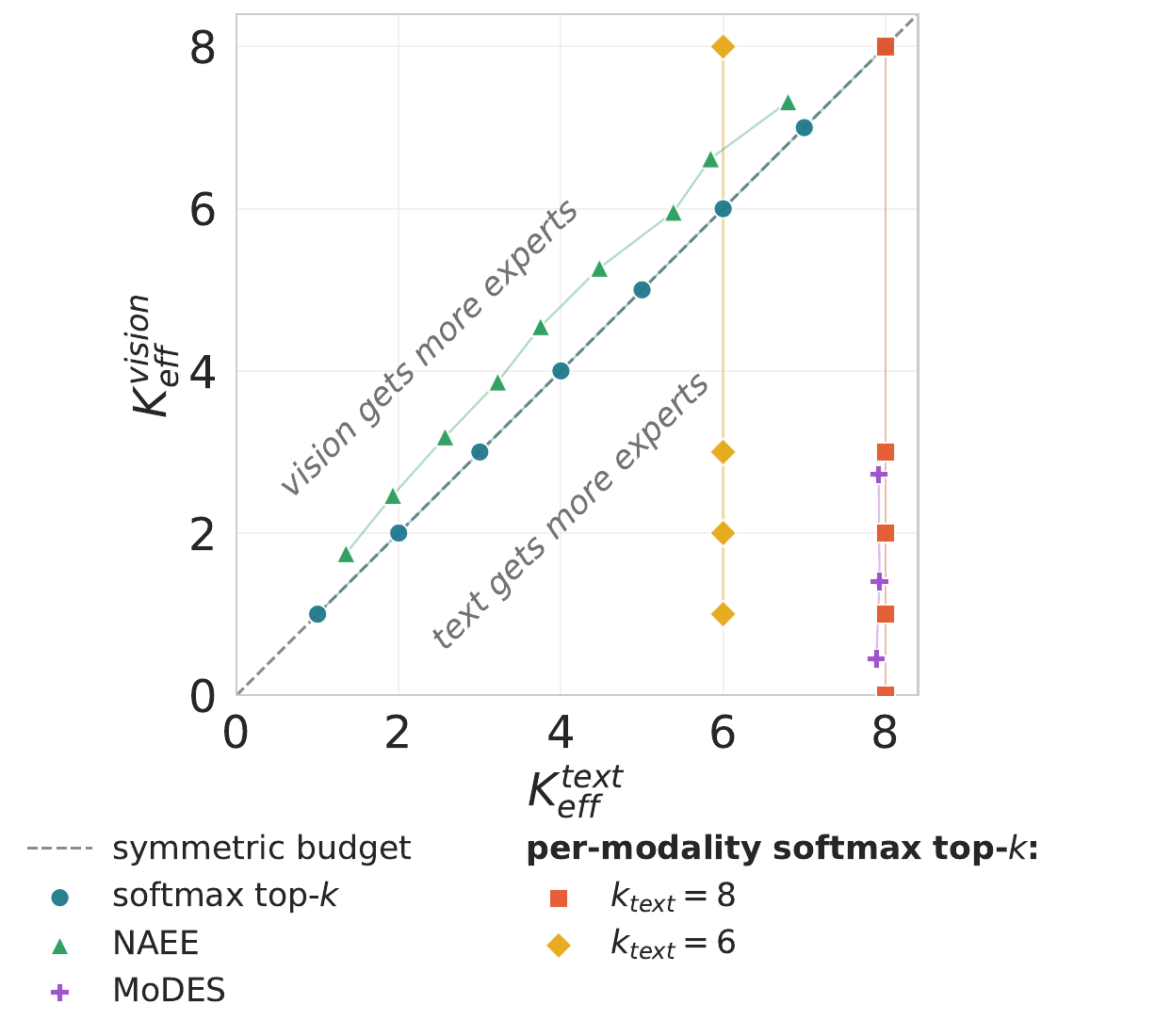}
    \caption{Measured \EK decomposed by modality, aggregated across the six multimodal benchmarks. The dashed diagonal marks a symmetric budget. The softmax \topk baseline lies on it by construction. NAEE traces the softmax \topk baseline, while MoDES produces a similar pattern to the per-modality softmax \topk with $k_\text{text}=8$.}
    \label{fig:effk_modality_differences}
\end{wrapfigure}

Figure~\ref{fig:effk_modality_differences} decomposes the measured \EK by modality, showing how each strategy distributes its expert budget. Softmax \topk and NAEE remain close to the symmetric diagonal, with NAEE allocating slightly more experts to vision tokens than to text tokens. This is the opposite of what the quality results favor. In contrast, MoDES retains nearly the full expert count for text tokens and concentrates skipping on vision tokens, closely following the per-modality baseline with $k_\text{text}=8$. This is in line with the quality--compute frontiers, highlighting similar performance for similar skip patterns.

MoDES does provide a benefit in the sub-one expert regime. Without
a finer (e.g.\ per-layer) granularity, our per-modality approach can
only reduce \EK{} below $1$ by setting $k_\text{vision}=0$, which
degrades performance to the point of limited practical use.

\subsubsection{Calibrated and measured \texorpdfstring{\EK}{Keff}}
\label{sec:nominal-and-measured}

The deviation between \textit{calibrated} and \textit{measured} \EK is also substantial in these benchmarks; Table~\ref{tab:nominal_vs_measured_keff} quantifies this gap for the MoDES results. This deviation stems from two main factors: the modality split and routing scores. Depending on the actual workload, sequences may have a different share of image vs. text tokens. Since more experts are skipped for image tokens, a lower proportion of image tokens shifts \EK upward, and vice versa. A second factor is differences in routing scores. Since MoDES uses the routing scores for skipping, different routing score distributions in the evaluation data may also influence how many experts are actually skipped. Both factors introduce a degree of unpredictability: the actual average number of remaining active experts can deviate from the calibrated figure by more than 50\%, even on benchmarks from the same task family as the calibration set (short-answer VQA; cf. Appendix~\ref{append:benchmarks}).

When evaluating on a text-only benchmark, in this case GSM8K~\citep{cobbe_training_2021}, the pattern is amplified. Text-only prompts are a typical use case for vision-enabled models, since many modern models bundle multimodal support into an otherwise general-purpose base model. Table~\ref{tab:nominal_vs_measured_keff} shows that for GSM8K, the \EK{} for MoDES shifts to 7.96, far from
the \EKn 1.2: the calibrated skipping effectively deactivates, and MoDES
retains quality only by reverting to nearly the full model: the calibrated
compute saving vanishes. Our per-modality baseline also shifts with the
workload, but is bounded by construction at \EK$=k_\text{text}=6$, where it
lands exactly, trading 2.7\,pp of accuracy (68.8\% vs.\ 71.5\%) for 25\% fewer
active experts than MoDES actually uses.
\begin{table}[t]
    \footnotesize
    \centering
    \caption{\EKn calibrated on GQA vs. measured \EK for MoDES on Qwen3-VL, averaged over six multimodal benchmarks. Text-only (GSM8K) is evaluated at the most aggressive skipping ratio. \EK shifts substantially based on the modality mix of the evaluation data.}
    \begin{tabular}{cccc}
        \toprule
        \multirow{2}{*}{MoDES Target Skip Ratio} & \multirow{2}{*}{\EKn} & \multicolumn{2}{c}{Measured \EK} \\
        \cmidrule(lr){3-4}
         & & Vision Benchmarks & Text-Only \\
        \midrule
        0.85 & 1.20  & 0.58 & 7.96 \\
        0.73 & 1.99 & 1.50 & ---  \\
        0.60 & 3.17  & 2.81 & ---  \\
        \bottomrule
    \end{tabular}

    \label{tab:nominal_vs_measured_keff}
\end{table}

\section{Discussion}
\label{sec:discussion}
While the trained or calibrated methods such as NAEE and the Allocator (Ada-K) can provide quality benefits compared to the baseline in some cases, these are in our experiments limited to $2.6$\,pp, often also in regimes where quality has already degraded substantially. Furthermore, the high dependence on calibration data suggests that any minor performance benefit from one of the calibrated strategies comes at the cost of performance in another area or domain. Beyond the standard calibration, identifying operating points and hyperparameters is a challenge, especially for RL-based Ada-K~\citep{eimer_hyperparameters_2023}.

The multimodal results support the same overall conclusions: MoDES outperforms both NAEE and the modality-unaware static baseline, showing that \emph{modality is valuable for deciding whether to skip experts} for a given token. However, MoDES's advantage over a calibration-free modality-aware baseline is confined to the sub-one \EK regime; for larger \EK the static baseline matches or outperforms it. We conclude that calibrated thresholds provide a benefit beyond treating vision and text tokens independently only in this sub-one \EK regime, but at the cost of complexity with only a modest runtime improvement as shown in Appendix~\ref{append:runtime_performance}.

Beyond the one-time calibration effort, matching strategies at equal \EK also
understates the true runtime cost. Every dynamic method adds a routing
decision to each forward pass. Dynamic skipping also forfeits the uniform per-token
compute of static \topk, and with it the load balancing and latency guarantees
quantified in Appendix~\ref{append:runtime_performance}. Both effects are excluded from our \EK-matched analysis, which
therefore favors the adaptive strategies. Finally, the
calibrated methods offer no guarantee on the realized \EK: as Table~\ref{tab:nominal_vs_measured_keff}
shows, \textit{measured} \EK can deviate substantially from the
\textit{calibrated} target.

\paragraph{Practical Recommendation} The added benefit of the dynamic skipping methods rarely outweighs the required calibration overhead and complexity. For a practical multimodal inference scenario, statically reducing $k_\text{vision}$ to 1--2 and $k_\text{text}$ to 6 for a model with eight active experts provides a strong trade-off between quality ($<5\%$ drop) and inference speed-up. For text-only models, the approach is similar: reducing $k$ to 6 provides a measurable inference speed-up with only a minor quality loss. Going beyond this point becomes heavily model-dependent and should be validated with a generative benchmark, as models tend to perform differently at those points.

To verify that this recommendation is not an artifact of the benchmarks used for its selection, we apply the operating point unchanged to MVBench \citep{li_mvbench_2024}, a held-out video benchmark. With $k_\text{vision}=1$
and $k_\text{text}=6$, the model achieves an accuracy of $69.95$
versus $71.40$
for the full model, a relative drop of $2.0\%$ at a measured \EK of 1.19. As video sequences are dominated by vision tokens, this represents an $85\%$ reduction in active experts. This minimal quality loss, on a workload and modality mix entirely outside the data used for calibration or selection, demonstrates the transferability of our recommendation. We also evaluate a larger model, GLM-4.6V \citep{vteam2025glm45vglm41vthinkingversatilemultimodal}, where the same operating point incurs an average drop of 4.2\% (Appendix~\ref{sec:scaling}).

\paragraph{Method design} Our results further highlight a significant gap in existing research: across compared methods as well as concurrent work such as ACE~\citep{xu_ace_2026} comparisons lack a consistent static-$k$ control at matched \emph{measured} compute and modality. When controlling for both, methods fail to consistently outperform the baseline. We therefore recommend that every future method compares against a static control at matched (measured) \EK that also uses any structural signal (e.g.\ modality) the method exploits.

\section{Conclusion}
Across three models as well as two held-out checks, none of the calibrated or learned skipping methods consistently outperform a static reduction of $k$ at matched measured \EK. The methods add calibration and tuning overhead and their realized \EK can deviate from the calibrated target ($\geq 6 \times$ in our experiments). This holds for text-only and multimodal models alike. We recommend static
per-modality $k$ as the default, and a static control at matched measured \EK
as the baseline for future methods.
\subsection*{AI use statement}
In this work, we used generative AI tools for implementing methods (including general coding assistance) and providing feedback on research methodology. We have not used generative AI tools for proposing or refining hypotheses, interpreting results, or developing theoretical models or conceptual frameworks. Generating synthetic datasets, formulating mathematical claims, providing critical ingredients for proving mathematical claims, writing proofs, translation, cleaning and reformatting datasets, and qualitative or thematic data analysis are not applicable to this work. Additionally, we used generative AI tools for literature research, brainstorming and editing for readability. We have reviewed all AI-assisted work. All LLM-generated code was reviewed by the authors and validated against existing results from literature. We take responsibility for the final content of this work, including text, claims or artifacts produced with the aid of generative AI.

\bibliography{iclr2027_conference}
\bibliographystyle{iclr2027_conference}
\pagebreak
\appendix
\section{Appendix}

\subsection{Taxonomy}
\label{app:taxonomy}
We provide the full decision rules in Table~\ref{tab:taxonomy}.

\newcolumntype{Y}[1]{>{\hsize=#1\hsize\raggedright\arraybackslash}X}

\begin{table*}[!h]
  \centering
  \footnotesize
  \renewcommand{\arraystretch}{1.15}
    \caption{Taxonomy of dynamic expert-skipping strategies, ordered by increasing
  signal complexity over the standard router. For NAEE we use the cumulative
  top-$k$ rule (App.~A.2 of \citet{lu_not_2024}); for DiEP we generalize the
  official top-2 rule to top-$k$ by comparing every selected expert against the
  dominant one, with the pairwise CKA factor $\gamma_2$ reimplemented from the
  paper (omitted in the public release). For quantile-calibrated methods we sweep the calibration quantile rather than fixing the paper's median. Approximate calibration time was measured with the
  official releases (where available) on Qwen3/Qwen3-VL on a single NVIDIA A100 80GB.}
  \begin{tabularx}{\textwidth}{@{}Y{1.0} Y{0.85} Y{1.25} Y{1.05} Y{0.85} l@{}}
    \toprule
    \textbf{Method} & \textbf{Criteria} & \textbf{Decision Rule} & \textbf{Calibration} & \textbf{Tunable} & \makecell[b]{\textbf{Calib.}\\\textbf{Time (h)}} \\
    \midrule
    Softmax top-$k$ (baseline)
      & Gate weights
      & Keep top-$k$ (fixed)
      & ---
      & $k$
      & --- \\
    Per-modality softmax top-$k$ (ours)
      & Gate weights
      & top-$k_\text{text}$ or top-$k_\text{vision}$ by modality (fixed)
      & ---
      & $k_\text{vision}$, $k_\text{text}$
      & --- \\
    NAEE \citep{lu_not_2024}
      & Gate weights (cumulative)
      & Keep top-$i^{\star}$, sorted desc.:\newline
        $i^{\star}=\min i$ \,s.t.\newline
        $\sum_{m>i} w_m \le \beta\sum_{m=1}^{k} w_m$
      & $\beta$: per-layer, calibrated
      & $\beta$
      & 0.5 \\
      DiEP \citep{bai_diep_2025}
      & Gate weights $+$ inter-expert CKA
      & Skip position $i\geq1$ if\newline
        $w_i < \gamma_1\,\gamma_2(e_0,e_i)\, w_0$
      & $\gamma_1$: per-layer quantile of $w_1/w_0$ (swept);\newline
        $\gamma_2$: pairwise expert CKA over layer mean
      & $\gamma_1$ quantile, $\gamma_2$
      & 0.5\\
    MoDES \citep{huang_modes_2026}
      & Gate prob.\ $\times$ global layer importance $\alpha^{(l)}$
      & Skip expert $i$ if $\alpha^{(l)}\pi_i < \tau_{\{t,v\}}$
      & $\alpha^{(l)}$: mean KL of full-layer skip
      & $\tau_t,\tau_v$
      & 20 \\
    Ada-K \citep{yue_ada-k_2025} (Allocator)
      & Token hidden state (learned)
      & Allocator $\rightarrow k^{\star}$; route top-$k^{\star}$
      & Trained (PPO), base frozen
      & alloc.\ weights, $\lambda$
      & 2.5 \\
    \bottomrule
  \end{tabularx}

  \label{tab:taxonomy}
\end{table*}
\subsection{Benchmark details}
\label{append:benchmarks}
We provide full details on the benchmarks conducted in Table~\ref{tab:benchmarks}.
\begin{table*}[!h]
  \centering
  \footnotesize
  \renewcommand{\arraystretch}{1.3}
  \setlength{\tabcolsep}{4pt}
    \caption{Overview of benchmarks with the corresponding calibration sets, evaluation
  metrics, and test-set sizes. For each calibration set, 1024 samples were used. The Allocator used up to 10000 samples from the same set, limited by split size. ``Mixed'' refers to our diverse pretraining dataset used for calibration. The
  multimodal strategies are calibrated on GQA following the MoDES
  protocol~\citep{huang_modes_2026}. Modality splits (text/vision token percentages) are approximated from 200 samples per dataset, tokenized with the Qwen3-VL processor, and reflect prefill only. }
  \begin{tabular}{@{}p{3.0cm} p{2.4cm} p{2.7cm} r p{1.6cm}@{}}
    \toprule
    \textbf{Benchmark} & \textbf{Calibration Sets} & \textbf{Metric} & \textbf{Test Samples} & \textbf{Split (text/vision)} \\
    \midrule
    ARC-Challenge \citep{clark_think_2018}
      & ARC-C Train, Mixed
      & \texttt{acc\_norm}
      & 1172 & --- \\
    HellaSwag \citep{zellers_hellaswag_2019}
      & HellaSwag Train, Mixed
      & \texttt{acc\_norm}
      & 10042 & --- \\
    GSM8K \citep{cobbe_training_2021}
      & GQA, Mixed
      & strict-match
      & 1319 & --- \\
    HumanEval \citep{chen_evaluating_2021}
      & Mixed
      & \texttt{pass@1}
      & 164 & --- \\
    ChartQA \citep{masry_chartqa_2022}
      & GQA \citep{hudson_gqa_2019}
      & relaxed accuracy
      & 2500 & 9.0/91.0 \\
    TextVQA \citep{singh_towards_2019}
      & GQA
      & VQA accuracy
      & 5000 & 2.4/97.6 \\
    MME \citep{fu_mme_2025}
      & GQA
      & total score (perception $+$ cognition)
      & 2374 & 3.4/96.6 \\
    DocVQA \citep{mathew_docvqa_2021}
      & GQA
      & ANLS
      & 5349 & 0.6/99.4\\

    InfoVQA \citep{mathew_infographicvqa_2021}
      & GQA
      & ANLS
      & 2801 & 0.6/99.4\\

    OCRBench \citep{liu_ocrbench_2024}
      & GQA
      & accuracy
      & 1000 & 15.7/84.3\\
    \bottomrule
  \end{tabular}

  \label{tab:benchmarks}
\end{table*}

\subsection{Calibration Sensitivity}
\label{app:calibration_sensitivity}
All considered methods except the static baselines require calibration data.
To establish the impact of the calibration data choice, we compare two regimes on
ARC-C and HellaSwag: \emph{best-case}, calibrated on the benchmark's own train
split, and \emph{realistic}, calibrated on a mixed corpus of English, Chinese, code,
and math built from \textit{C4\_en} and \textit{C4\_zh} \citep{raffel_exploring_2023},
\textit{open-web-math} \citep{paster_openwebmath_2023}, and
\textit{python-codes-25k} \citep{flytech_python_codes_25k}. The realistic regime
simulates the calibration--evaluation mismatch typical of actual deployments, while
the best-case regime represents the most favorable calibration setting.

\begin{figure}
    \centering
    \includegraphics[width=\textwidth]{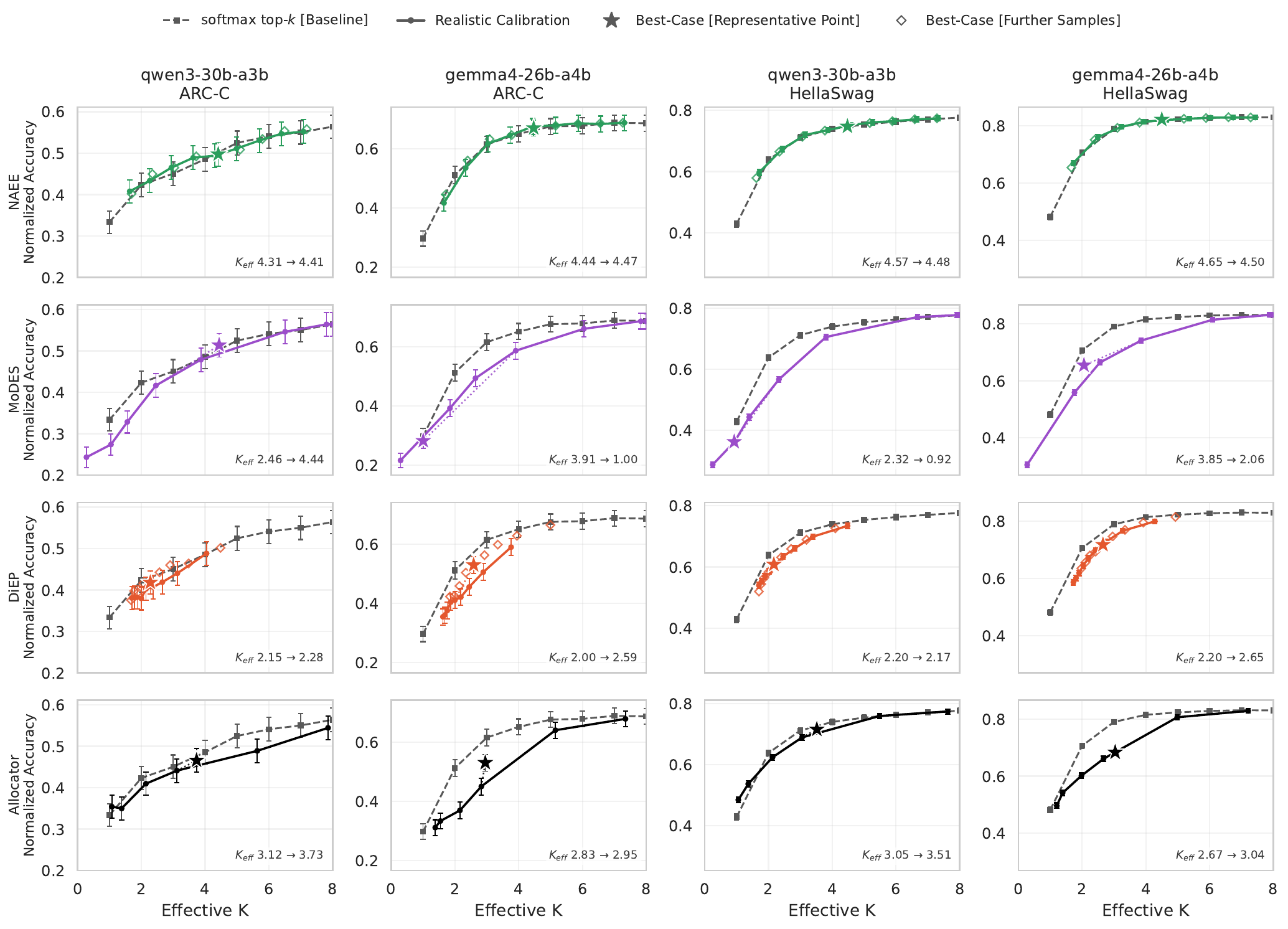}
    \caption{Comparison of best-case and realistic calibration for Qwen3 and Gemma4 across ARC-C and HellaSwag. We select the middle threshold/quantile/cost parameter of the realistic calibration as our representative point. Error bars show the $95\%$ Confidence Interval (CI) from lm-eval standard errors. Any shift in \EK{} (as for example observed for MoDES) is due to the change in calibration data alone. Panel annotations provide this shift in realized \EK, realistic $\rightarrow$ best-case. Even in a best-case calibration scenario the dynamic skipping strategies do not consistently outperform the baseline.}
    \label{fig:calibration-sensitivity}
\end{figure}

Figure~\ref{fig:calibration-sensitivity} highlights the impact of the calibration data. We select the middle point of the corresponding threshold sweep and observe the change in quality between best-case and realistic calibration. In some cases this best-case calibration does provide benefits compared to the realistic calibration. However, these improvements are marginal and do not materially affect the performance relative to the baseline. Further, we observe that the change in calibration set leads to substantial shifts in \EK at identical thresholds. NAEE, the best-performing dynamic skipping method in the practically relevant regime, is also the most robust to the choice of calibration data, showing only minimal shifts under best-case calibration. Preliminary experiments highlighted this as well: even a mismatched calibration set (MATH only) led to very similar performance.

\subsection{Allocator Training}
\label{append:Allocator_training}
We provide the full set of training parameters for our Allocator in Table~\ref{tab:allocator-training-config-ppo}.

\begin{table}[t]
  \centering
  \caption{Training configuration for the learned allocator trained with the PPO
    objective (mixed calibration). The same configuration is used for Qwen3-30B-A3B and
    Gemma4-26B-A4B; $\lambda$ enters the PPO loss, so each point of the cost-penalty
    sweep is a separately trained allocator.}
  \label{tab:allocator-training-config-ppo}
  \begin{tabular}{ll}
    \toprule
    \textbf{Hyperparameter} & \textbf{Value} \\
    \midrule
    Objective                    & PPO \\
    Sample-based evaluation      & enabled \\
    Calibration dataset          & \texttt{mixed} \\
    Calibration samples          & 10000 \\
    Max sequence length (tokens) & 16384 \\
    Training steps               & 312 \\
    Micro-batch size             & 1 \\
    Gradient accumulation steps  & 32 \\
    Effective batch size         & 32 \\
    Sequence length (training)   & 2048 \\
    Learning rate                & $1\times10^{-3}$ \\
    Warmup fraction              & 0.05 \\
    Final LR fraction            & 0.1 \\
    KL temperature               & 1.0 \\
    $K_{\max}$ multiplier        & 16 \\
    Validation set size          & 128 \\
    Validation interval (steps)  & 50 \\
    Random seed                  & 1234567 \\
    \midrule
    \multicolumn{2}{l}{\textit{PPO-specific}} \\
    Discount factor $\gamma$     & 0.99 \\
    Clip range $\epsilon$        & 0.2 \\
    PPO epochs                   & 2 \\
    Entropy coefficient          & 0.0 \\
    Advantage whitening          & disabled \\
    \midrule
    \multicolumn{2}{l}{\textit{P-Warm warm-start}} \\
    Policy warmup                & enabled \\
    Warm-start steps             & 312 (full length) \\
    $p^{*}$ search grid          & 18 points, linear in $[0.1,\,0.95]$ \\
    \midrule
    \multicolumn{2}{l}{\textit{Optimization (code defaults; not recorded per run)}} \\
    Optimizer                    & AdamW ($\epsilon=10^{-6}$) \\
    Weight decay                 & 0.1 \\
    LR schedule                  & cosine (warm-start), constant (PPO) \\
    Warm-start learning rate     & $1\times10^{-3}$ \\
    \midrule
    Cost penalty sweep ($\lambda$) & $\{0.0003,\,0.001,\,0.005,\,0.01,\,0.03,\,0.08\}$ \\
    \bottomrule
  \end{tabular}
\end{table}

\subsection{Runtime Performance}
\label{append:runtime_performance}
To validate \EK as a proxy for inference performance, we measure end-to-end inference with statically constrained expert counts using vLLM \citep{kwon_efficient_2023}. This constitutes an upper bound on the gains achievable by any strategy at a given \EK, as it excludes both routing-decision overhead and raggedness in per-token expert allocation; the static \topk baseline attains this bound by construction. Standardized evaluation uses GuideLLM \citep{noauthor_guidellm_2024} to benchmark Qwen3-30B-A3B on a single NVIDIA A100 80GB GPU, in two scenarios: synchronous (batch size 1, memory-bandwidth bound) and throughput (batch size 256, compute bound). We use GuideLLM's `chat' scenario to generate prompts. Figure~\ref{fig:eff_k_vs_perf} reports time-to-first-token (TTFT), inter-token latency (ITL), and output throughput. Reducing \EK from 8 to 1 lowers TTFT and ITL by 18\% in the synchronous scenario and lowers TTFT by $\sim30\%$ and ITL by $\sim25\%$ in the throughput scenario, while output throughput rises from 111 to 135 tokens/s ($1.22\times$) and from $\sim1.9$k to $\sim2.65$k tokens/s ($1.39\times$), respectively. Notably, these gains are far below the $8\times$ reduction in expert compute: even under ideal static skipping, the end-to-end benefit is bounded well below proportionality, further compressing the value of the marginal quality differences between skipping strategies. The MoE blocks of the model only make up part of the overhead incurred; components such as attention, KV-Cache traffic, and non-expert weights are not addressed by any skipping technique and still incur computational costs.

\subsection{Detailed Benchmark Results}
\label{append:benchmark_results}
\begin{figure}[ht]
    \centering
    \includegraphics[width=\linewidth]{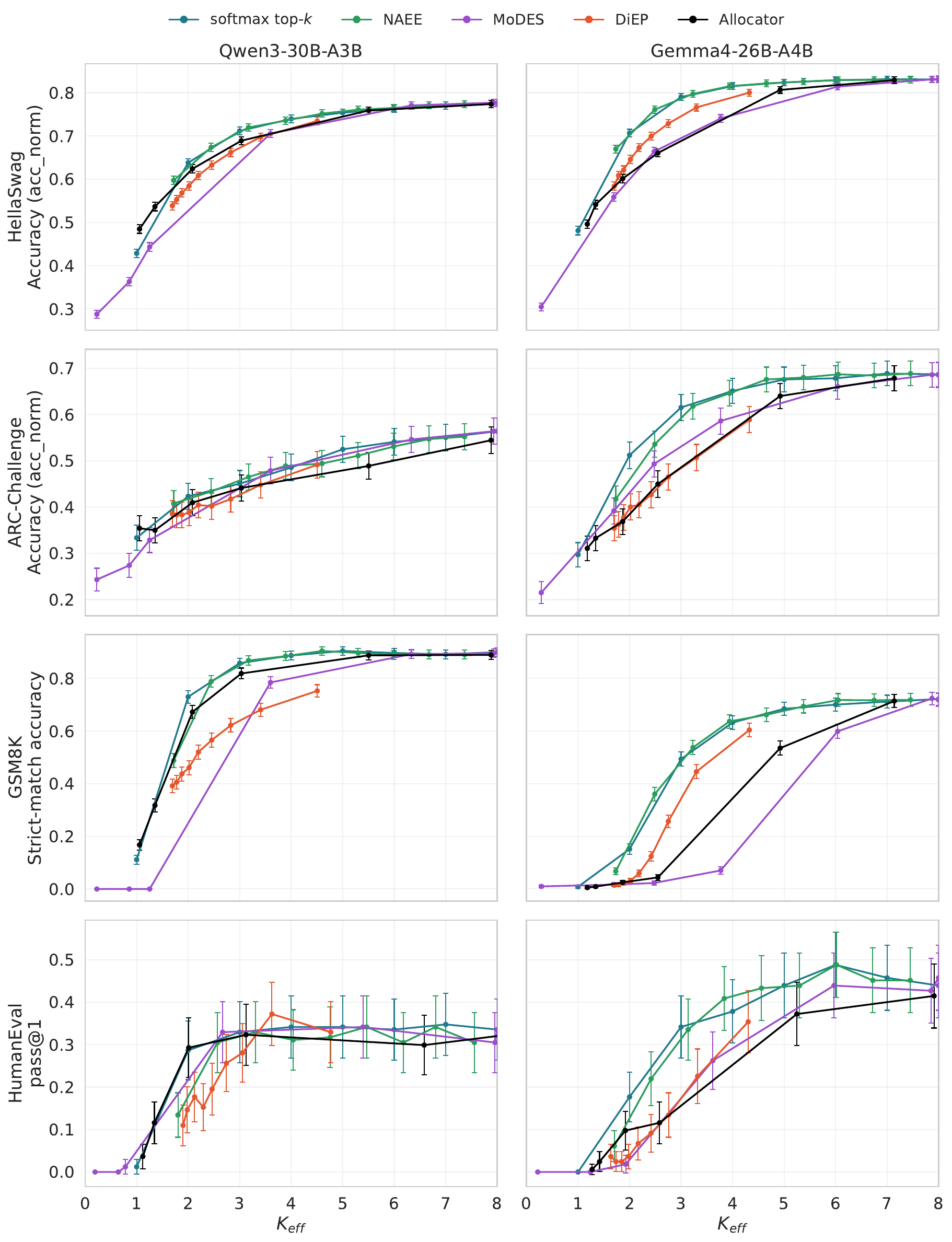}
    \caption{Comparison of model performance with different expert skipping strategies. Error bars show the $95\%$ confidence interval calculated based on the lm-eval standard error. For ARC-C and HellaSwag the results are all relatively close, while for GSM8K and HumanEval we see a more clear separation.}
    \label{fig:absolute_metrics_text_stderr}
\end{figure}

\begin{figure}[ht]
    \centering
    \includegraphics[width=\linewidth]{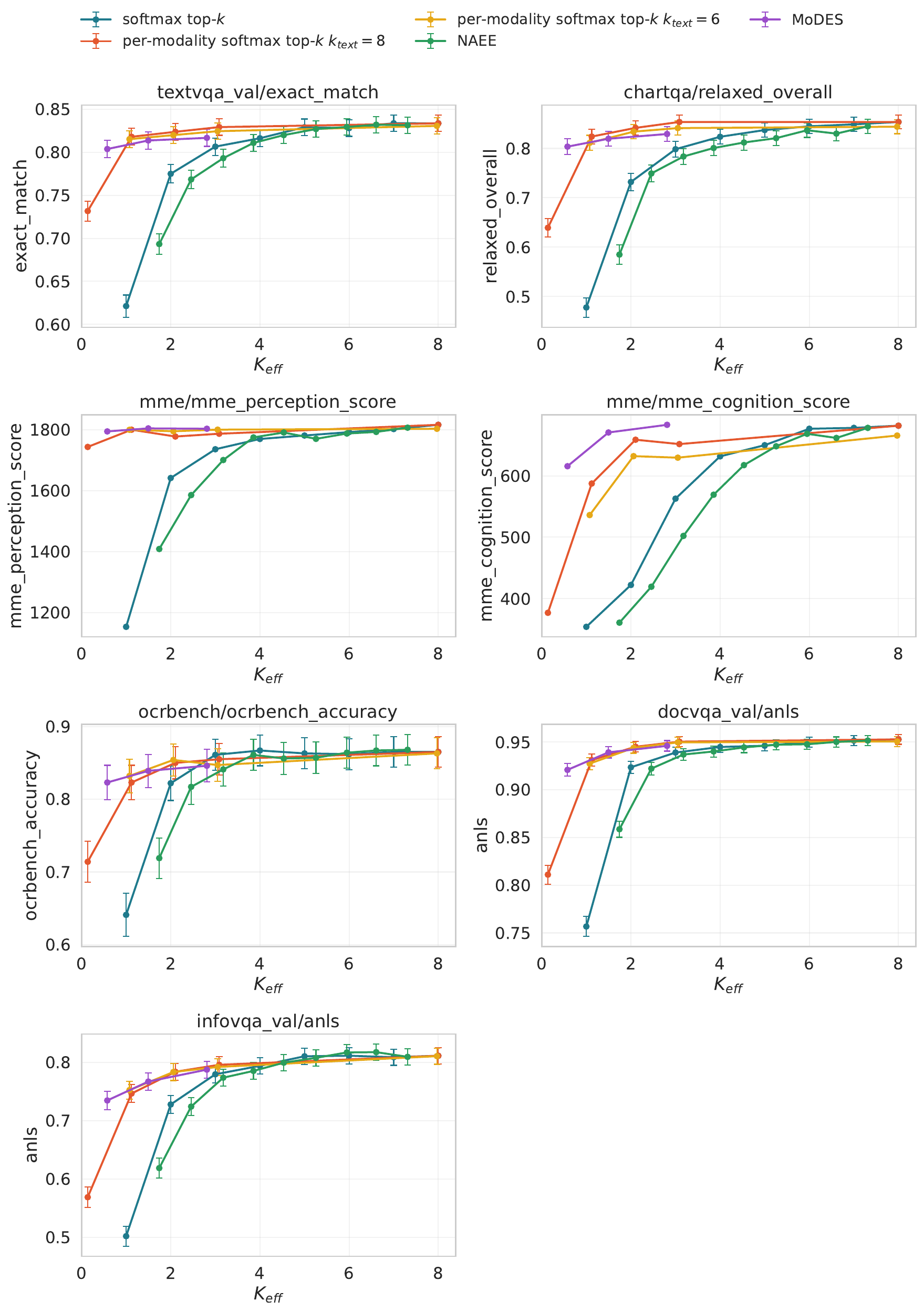}
    \caption{Comparison of model performance with different expert skipping strategies. MME Cognition Score is an outlier, where MoDES shows clear benefits compared to the other strategies. The $95\%$ CI is provided for TextVQA and ChartQA based on the standard error reported by LMMS-Eval. For MME this is not included due to the non-linear nature of the score.}
    \label{fig:absolute_metrics_multimodal_stderr}
\end{figure}

To highlight the differences in performance we provide the detailed per-benchmark sweep results in Figures~\ref{fig:absolute_metrics_text_stderr} and~\ref{fig:absolute_metrics_multimodal_stderr}. A number of differences can be observed in both the text and multimodal context.

\paragraph{Text-only} In the text-only results, the performance characteristics for our generative benchmark and the multiple-choice benchmarks differ. ARC-C and HellaSwag performance decays smoothly towards the random selection floor, with all methods performing reasonably close to one another. For GSM8K this changes dramatically: for both models, the drop towards zero score is apparent and linked to dropping the number of experts beyond a certain point. Especially for Gemma4, no strategy is able to retain reasonable performance once fewer than 3 experts remain active, and some strategies already fail at 4. This is likely a combination of factors: first, we use strict-match for evaluation. This means that the model will score 0 when it fails to match the correct output format. Second, MoDES in particular can reduce the number of experts to 0 for some tokens, potentially breaking causal generation or reasoning. HumanEval performs similar to GSM8K.

\paragraph{Multimodal} For multimodal evaluation the performance varies less across benchmarks with a notable exception. The MME perception score shows limited drop even when vision tokens receive zero experts, provided enough experts are allocated to the text tokens. This differs from the MME cognition score, where the drop is more apparent and MoDES markedly outperforms the baseline rather than simply matching it.


\subsection{Validating Implementations}
\label{append:validating_implementations}
Due to the incomplete nature of some of the code releases, we validate each method in the setting reported by its original authors.

\paragraph{NAEE}
The NAEE code release only contains implementations for Mixtral 8x7B. To validate our implementation, we aim to reproduce their domain-specific calibration and evaluation regime. We calibrate Mixtral 8x7B \citep{jiang_mixtral_2024} on the MATH dataset \citep{hendrycks_math_2021} and evaluate performance on 5-shot GSM8K \citep{cobbe_training_2021} using strict-match accuracy.  We use NAEE's default $q=0.5$ quantile (median) skip criterion, as described in their methodology. The reported accuracy from the NAEE paper is $54.28$ \citep{lu_not_2024}, which corresponds to a $7.4\%$ relative drop compared to the original $58.61$. Our reimplementation achieved a baseline score of $58.52$ and an expert skipping score of $55.42$ (a 5.3\% relative drop). This minor difference could be attributable to a number of factors, including slight differences between the cumulative-rule from the appendix and the two expert rule they describe in the paper. However, as the scores are very close, we are confident our implementation does not disadvantage NAEE in the comparison.

\paragraph{DiEP}
Validation on DiEP was conducted on Mixtral 8x7B, evaluating the average accuracy across MMLU~\citep{hendrycks_measuring_2021b}, OpenBookQA~\citep{mihaylov_can_2018}, BoolQ~\citep{clark_boolq_2019}, and RTE~\citep{dagan_pascal_2005}. Calibration was done on the ``mixed'' set, as it has significant overlap with the original C4~\citep{raffel_exploring_2023} calibration data used in the DiEP paper~\citep{bai_diep_2025}. Since the complete skip configuration is not explicitly reported, we again use the $q=0.5$ quantile calibration that serves as their operating point elsewhere in the paper. \citet{bai_diep_2025} report an average accuracy of $65.1$ for the base model and $64.1$ after skipping, a $1.6\%$ relative drop; per-task scores are reported for the base model only. As Table~\ref{tab:diep_validation} shows, our measured baseline matches theirs on every task when OpenBookQA is scored with unnormalized accuracy ($65.6$ vs.\ $65.1$ on average); the apparent baseline gap under our default reporting ($68.6$) stems entirely from using length-normalized accuracy on OpenBookQA ($49.0$ vs.\ $37.2$ unnormalized). Our reimplementation, which includes the from-scratch CKA computation absent from the public release, yields a relative drop of $2.6\%$ in DiEP's metric convention ($65.6 \rightarrow 64.0$; $2.1\%$ with length-normalized OpenBookQA). The difference to the reported $1.6\%$ is dominated by RTE ($71.5 \rightarrow 67.5$), the smallest of the four tasks ($277$ test examples, standard error $\approx 2.7$\,pp), and is therefore within evaluation noise; the remaining tasks degrade by at most $1.4$ points. We conclude that our implementation reproduces the reported behavior closely.

\begin{table}[h]
  \centering
    \caption{Per-task validation of our DiEP reimplementation on Mixtral 8x7B. \citet{bai_diep_2025} report per-task scores for the base model and only the aggregate after skipping ($64.1$). Our baseline matches theirs on all tasks under their metric convention; the aggregate difference in our default reporting is explained by length-normalized accuracy on OpenBookQA.}
  \small
  \renewcommand{\arraystretch}{1.2}
  \begin{tabular}{@{}lccc@{}}
    \toprule
    Task & DiEP~\citep{bai_diep_2025} (base) & \makecell{Reimplementation\\(Base)} & \makecell{Reimplementation\\(skipping)} \\
    \midrule
    MMLU (acc)               & 67.9 & 68.0 & 67.0 \\
    OpenBookQA (acc)         & 35.4 & 37.2 & 35.8 \\
    OpenBookQA (acc\_norm)   & ---  & 49.0 & 48.6 \\
    BoolQ (acc)              & 85.3 & 85.9 & 85.5 \\
    RTE (acc)                & 71.5 & 71.5 & 67.5 \\
    \midrule
    Average (OBQA acc)       & 65.1 & 65.6 & 64.0 \\
    Average (OBQA acc\_norm) & ---  & 68.6 & 67.1 \\
    \bottomrule
  \end{tabular}

  \label{tab:diep_validation}
\end{table}

\paragraph{MoDES}
For multimodal MoDES we use the original calibration method included with their code release. As we reimplement the routing function, we validate it by comparing our results against the reported values for TextVQA in Table~\ref{tab:modes_validation}. We use the same routing function for the text-only evaluation, but deviate from the original calibration: we calibrate on the text datasets and conduct a sweep across thresholds rather than using the frontier search described in the paper. Since the text-only evaluation uses the identical routing function, validated above, and differs only in the calibration procedure, we treat it as a faithful application of the method rather than a separate reproduction.

\begin{table}[h]
  \centering
    \caption{MoDES reproduction on TextVQA (exact-match accuracy), evaluated at the paper's exact calibration target ratios $\rho$. As both use identical thresholds, the comparison is at matched operating points; our reimplementation reproduces the reported results within 0.6 points across the range.}
  \small
  \renewcommand{\arraystretch}{1.2}
  \begin{tabular}{@{}ccc@{}}
    \toprule
    Target ratio $\rho$ & MoDES~\citep{huang_modes_2026} & Reimplementation \\
    \midrule
    0.60 & 81.82 & 81.68 \\
    0.73 & 81.65 & 81.38 \\
    0.85 & 80.97 & 80.38 \\
    \bottomrule
  \end{tabular}

  \label{tab:modes_validation}
\end{table}

\paragraph{Ada-K}
No public implementation or model checkpoints were available for Ada-K as of writing. As such, we reproduced its classifier architecture
ourselves, which we refer to throughout as the Allocator. We train it using the same PPO approach as the original Ada-K paper. Some variance is nonetheless expected, due to the nature of reinforcement learning: its sensitivity to hyper-parameters \citep{eimer_hyperparameters_2023} is itself part
of the practical complexity we discuss in Section~\ref{sec:discussion}.
To validate the performance of our Allocator, we ran Qwen1.5-MoE-A2.7B, trained the Allocator on our mixed dataset, and evaluated on 25-shot ARC-C. Our Allocator achieved a score of $53.92$ at \EK$=2.45$, very close to the reported $54.41$ at a slightly higher \EK$=2.58$, consistent with the slightly larger expert budget.

\subsection{Scaling to larger models}
\label{sec:scaling}
We also evaluate our recommendation on another (larger) model: GLM-4.6V \citep{vteam2025glm45vglm41vthinkingversatilemultimodal}. It natively has $k=8$ and 128 available experts. It also has a shared expert, differentiating it from Qwen3-VL. We evaluate the operating point $k_\text{text}=6$ and $k_\text{vision}=1$ and observe an average drop of 4.2\% compared to the original model across ChartQA, TextVQA and MME. Full results are included in Table~\ref{tab:multimodal_routing}. As can be seen the drop is dominated by MME-Cognition.

\begin{table}[t]
  \centering
  \caption{Comparison of the full model (top-$8$) and modality-aware routing
  ($k_\text{text}=6$, $k_\text{vision}=1$) on GLM-4.6V. Mean active experts per token
  for the routed model is $1.18$ across the benchmarks.}
  \begin{tabular}{llrrr}
    \toprule
    Benchmark & Metric & Baseline & Routed & $\Delta$ \\
    \midrule
    TextVQA (val) & Exact match (\%) & 83.57 & 82.16 & $-1.41$ \\
    ChartQA & Relaxed overall (\%) & 74.92 & 70.96 & $-3.96$ \\
    ChartQA & Relaxed human (\%) & 72.88 & 67.92 & $-4.96$ \\
    ChartQA & Relaxed augmented (\%) & 76.96 & 74.00 & $-2.96$ \\
    MME & Perception score & 1714.0 & 1671.9 & $-42.1$ \\
    MME & Cognition score & 671.8 & 579.3 & $-92.5$ \\
    \bottomrule
  \end{tabular}

  \label{tab:multimodal_routing}
\end{table}

\end{document}